\documentclass[10pt,twocolumn,letterpaper]{article}

\usepackage[pagenumbers]{cvpr}
\expandafter\ifx\csname iflatexml\endcsname\relax\expandafter\newif\csname iflatexml\endcsname\fi

\usepackage{array}
\usepackage{multirow}

\newif\ifverifydraft
\definecolor{verifyblue}{rgb}{0.05,0.30,0.75}

\newcommand{\V}[1]{\ifverifydraft{\color{verifyblue}#1}\else#1\fi}

\usepackage{microtype}

\DeclareMathOperator*{\argmax}{arg\,max}

\DeclareMathOperator{\med}{med}
\DeclareMathOperator{\IQR}{IQR}
\DeclareMathOperator{\clip}{clip}
\DeclareMathOperator{\ReLU}{ReLU}

\verifydraftfalse%

\newcommand{\srccMain}{\V{0.6739}}%
\newcommand{\srccGemini}{\V{0.6405}}%
\newcommand{\srccZCFD}{\V{0.5415}}%
\newcommand{\srccDreamSim}{\V{0.5364}}%
\newcommand{\srccOursGT}{\V{0.6462}}%
\newcommand{\ceiling}{\V{0.571}}%

\newcommand{\srccMORfour}{\V{0.7432}}%

\newcommand{\srccT}{\V{0.6738}}%
\newcommand{\speedup}{\V{nine}}%

\newcommand{\nCases}{808}

\newcommand{\nRatersAll}{117}%
\newcommand{\nRankingsAll}{1{,}873}%
\newcommand{\nBaselines}{29}

\definecolor{cvprblue}{rgb}{0.21,0.49,0.74}
\usepackage[pagebackref,breaklinks,colorlinks,allcolors=cvprblue]{hyperref}
\hypersetup{pdftitle={ERASCORE: Reference-Free Evaluation of Scene-Consistent Object Removal},pdfauthor={Yoojeong Lee, Sehwa Lee, Seogkyu Jeon, Kibeom Hong}}
\def\confName{CVPR}
\def\confYear{2027}

\title{ERASCORE: Reference-Free Evaluation of Scene-Consistent Object Removal}

\iflatexml
\author{
Yoojeong Lee\\
Sookmyung Women's University\\
\texttt{uujeong59@sookmyung.ac.kr}
\and
Sehwa Lee\\
Sookmyung Women's University\\
\texttt{1ee3flowers@gmail.com}
\and
Seogkyu Jeon\\
\texttt{jone9312@gmail.com}
\and
Kibeom Hong$^{\dagger}$\\
Sookmyung Women's University\\
\texttt{kb.hong@sookmyung.ac.kr}\\
$^{\dagger}$Corresponding author
}
\else
\author{Yoojeong Lee$^{1}$ \qquad Sehwa Lee$^{1}$ \qquad Seogkyu Jeon \qquad Kibeom Hong$^{1}$\thanks{Corresponding author.}\\
$^{1}$Sookmyung Women's University, Seoul, Korea\\
{\tt\small uujeong59@sookmyung.ac.kr, 1ee3flowers@gmail.com}\\
{\tt\small jone9312@gmail.com, kb.hong@sookmyung.ac.kr}
}
\fi

\begin{document}
\maketitle
\begin{abstract}
Evaluating object removal is challenging because ground-truth images are difficult to obtain and multiple fills can be valid for the same mask. Existing reference-free metrics construct contextual references from the removal result, restrict them to the mask neighborhood, or pool scene elements into a summary. We argue that a reliable reference should come from the original image, cover the entire unmasked scene, and keep scene elements separate. Following these principles, ERASCORE constructs patch-level Global Scene Keys from the original image and matches each removed-region patch to its nearest key independently in semantic and perceptual feature spaces. A geometry-aware hallucination term further checks object-like structures against scene regions and depth. ERASCORE requires no clean reference image, task-specific training, or weights fitted to human judgments. On 808 image-mask pairs from three benchmarks ranked by 117 participants, ERASCORE agrees with human rankings more closely than all 29 baselines, including GPT-4o and Gemini-2.5-Pro judges, achieving a Spearman correlation of 0.6739 and leading on every benchmark. Controlled ablations show that each reference principle contributes on its own. A lightweight variant runs about nine times faster while maintaining nearly the same agreement with human rankings.
\end{abstract}

\begin{figure}[t]
\centering
\includegraphics[width=\linewidth]{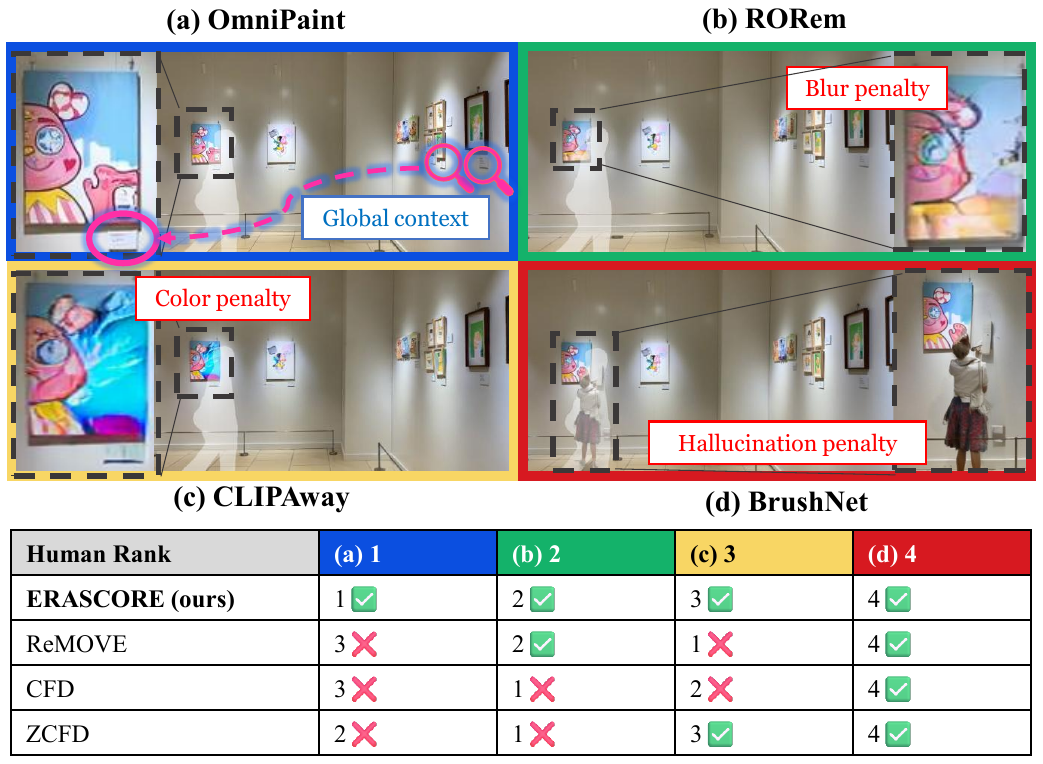}
\caption{\textbf{What each reference-free metric rewards.} Four removals of the same visitor. In (a), the label in the fill is supported by labels elsewhere in the gallery, which only a scene-wide reference can reach. The other outputs blur the fill (b), shift its color (c), or insert a new person (d). The table gives the rank each metric assigns among these four outputs, and the top row ranks them by the mean rank of four raters. A check mark denotes a rank that matches the human rank. Only ERASCORE reproduces the human order across the clean fill, blurred fill, color shift, and hallucination.}
\label{fig:teaser}
\end{figure}
\section{Introduction}
\label{sec:intro}

Judging whether an object removal succeeded is hard because the scene behind the target is occluded in the original image. A removal method erases a target and fills the hole so that the scene continues as if the target had never been there. Generative inpainting has made such fills look convincing~\cite{suvorov2022resolution,rombach2022high,zhuang2024task,winter2024objectdrop,ekin2024clipaway,yu2025omnipaint}, but a convincing fill is not necessarily a correct one. A result can continue the wrong surface, smear the texture it should restore, or place a new object where the target stood, and existing metrics often fail to tell these failures from success. Full-reference metrics~\cite{wang2004image,zhang2018unreasonable,fu2023dreamsim} require a ground-truth image of the same scene without the target. Such images are difficult to obtain for real scenes, and even a captured ground truth represents only one of many valid completions~\cite{li2026prove}. No-reference quality metrics~\cite{yang2022maniqa,ke2021musiq} require no such image, but they measure how natural the output looks, and a well-rendered hallucination can also look natural.

The evidence that these metrics lack is often already in the image. Most of the scene survives the edit, and the walls, floors, and foliage that continue behind a target frequently appear elsewhere in the frame as well. Reference-free metrics exploit this by judging the removed region against the visible context~\cite{chandrasekar2024remove,yu2025omnipaint,lee2026zcfd,li2026prove}. These designs differ mainly in how they build the reference, and three choices in that construction, its \emph{source}, \emph{scope}, and \emph{representation}, determine which evidence a metric can use. ReMOVE, CFD, and PROVE take the removal result as the source, so a method that alters the surroundings also alters the evidence used to judge it. Limiting the scope to a local crop or a box around the mask keeps a matching surface elsewhere in the image out of reach, and a correct fill is then penalized for lacking support that the scene does contain (Fig.~\ref{fig:teaser}a). A representation that pools features blends floor, wall, and sky into one vector that resembles none of them, and one that compares feature distributions does not record which scene element supports each patch of the removed region.

ERASCORE makes the opposite choice in each case. It reads every reference from the original image, so content generated by the removal method cannot directly supply its own reference evidence. Its reference covers the whole unmasked scene, and it stores scene elements separately, as patch-level Global Scene Keys and region-level Scene Region Keys, so that each patch of the removed region is compared with the element it most resembles, wherever that element lies. Three terms evaluate this evidence. A semantic term $S$ measures whether the fill belongs to something the scene contains, and a perceptual term $P$ measures whether it is rendered like real material in the scene, which exposes the blur and seams that semantic features largely tolerate. A geometry-aware hallucination term $H$ examines object-like regions inside the mask as wholes, since a coherent structure can pass patch-level tests while standing in front of the surface that should continue behind the target. The three terms are multiplied, with no weight fitted to human rankings.

ERASCORE agrees with human judgment more closely than every metric we tested. Against \nBaselines{} baselines spanning full-reference, no-reference, and reference-free metrics and multimodal LLM judges, it achieves the highest agreement with human rankings of \nCases{} samples from RORD~\cite{sagong2022rord}, OBER-Test, and OBER-Wild~\cite{zhao2026objectclear}, on every benchmark and overall, and it predicts a held-out rater's ranking better than another rater does. Controlled ablations trace the gain to each reference principle separately, and analyses of natural and constructed hallucinations show where region-level reasoning is needed.

Our contributions are as follows.
\begin{itemize}
\item We identify reference construction, namely its source, scope, and representation, as a design choice that distinguishes existing reference-free removal metrics, and we isolate each choice in a controlled ablation.
\item We propose ERASCORE, which retrieves support from the full original scene in separate semantic and perceptual spaces and checks object-like regions with scene-region and depth evidence, without task-specific training or fitted weights. A lightweight variant, ERASCORE-T, replaces the region branch with a target-residual test on features the metric already computes.
\item We collect \nRankingsAll{} rankings of the outputs of eight removal methods from \nRatersAll{} participants, with four independent rankings on 355 randomly assigned samples, and show that ERASCORE agrees with them better than all \nBaselines{} baselines.
\end{itemize}

\section{Related Work}
\label{sec:related}

\subsection{Object Removal}
\label{sec:related_removal}

Object removal is commonly cast as mask-guided inpainting, from the fast Fourier convolutions of LaMa~\cite{suvorov2022resolution} to latent diffusion~\cite{rombach2022high}, which PowerPaint~\cite{zhuang2024task} and BrushNet~\cite{Ju2024BrushNetAP} adapt with task prompts and a dual-branch design. Such inpainting priors can place a new object in the hole instead of continuing the background~\cite{ekin2024clipaway,Li2025RORemTA}. Removal-specific methods counter this with pairs that show a scene with and without the target, captured photographically by ObjectDrop~\cite{winter2024objectdrop}, mined from fixed-viewpoint video by RORD~\cite{sagong2022rord} and OmniEraser~\cite{Wei2025OmniEraserRO}, filtered with humans in the loop by RORem~\cite{Li2025RORemTA}, and extended with masks for object effects by ObjectClear~\cite{zhao2026objectclear}. Without paired data, CLIPAway~\cite{ekin2024clipaway} conditions on background-focused CLIP~\cite{radford2021learning} embeddings, OmniPaint~\cite{yu2025omnipaint} couples removal with insertion, and YOEO~\cite{YOEO} suppresses unexpected content.

\subsection{Evaluating Object Removal}
\label{sec:related_metrics}

\paragraph{Full-reference and no-reference metrics.}
Full-reference metrics such as PSNR, SSIM~\cite{wang2004image}, LPIPS~\cite{zhang2018unreasonable}, and DreamSim~\cite{fu2023dreamsim} need an image of the scene without the target, and a single captured proxy can penalize other valid completions~\cite{li2026prove}. No-reference quality metrics~\cite{yang2022maniqa,ke2021musiq} drop the reference but measure generic naturalness, which a well-rendered hallucination also satisfies.

\paragraph{Reference-free removal metrics.}
Reference-free removal metrics judge a removal from the input image, the mask, and the result, and they differ mainly in how they build the reference. ReMOVE~\cite{chandrasekar2024remove} compares mean encoder features of the Segment Anything Model (SAM)~\cite{kirillov2023segany} inside and outside the mask within a crop of the result. CFD (context-aware feature deviation)~\cite{yu2025omnipaint} compares DINOv2 features~\cite{oquab2024dinov2} of the removed region with the rest of a box around the mask and penalizes SAM proposals inside the mask that differ from adjacent proposals. The spatial metric of PROVE~\cite{li2026prove}, RC-S, compares feature distributions inside and around the mask with a sliding-window maximum mean discrepancy. ZCFD~\cite{lee2026zcfd} restricts the comparison to a depth-derived Z-layer, the surface expected to continue behind the target. In each case the evidence is read from the result, stays near the target or on one surface, or is pooled before it is compared. ERASCORE changes all three properties and uses the Z-layer of ZCFD only as a geometric prior. CrossScore~\cite{wang2024crossscore} shares the broader idea of scoring an image against other observations of the same scene, in a multi-view setting.

\paragraph{Learned and self-consistency evaluators.}
Learned evaluators include PAL4Inpaint~\cite{zhang2022pal4inpaint}, which segments perceptual artifacts in inpainted images and scores the share of the fill marked as artifact, and RORem~\cite{Li2025RORemTA}, which trains a discriminator on human feedback to select successful removals. Both depend on the failure modes in their training data. Re-inpainting self-consistency~\cite{chen2024reinpainting} avoids training but ties the score to a second inpainting model. ERASCORE needs neither training on removal data nor an auxiliary generator.

\paragraph{Multimodal LLM judges.}
Multimodal large language models (MLLMs) prompted as judges~\cite{ku2024viescore,openai2024gpt4o,comanici2025gemini} approach human agreement on several image synthesis tasks, though less reliably on editing than on generation~\cite{ku2024viescore}, and they now score editing benchmarks~\cite{Ye2025ImgEditAU,Jiang2026UniEditBenchAU}. They score on a few discrete levels, so competing methods often tie, and each call adds cost. We include GPT-4o and Gemini-2.5-Pro as listwise judges, and they are the strongest baselines in our study.

\section{Method}
\label{sec:method}

\begin{figure*}[t]
\centering
\includegraphics[width=\textwidth]{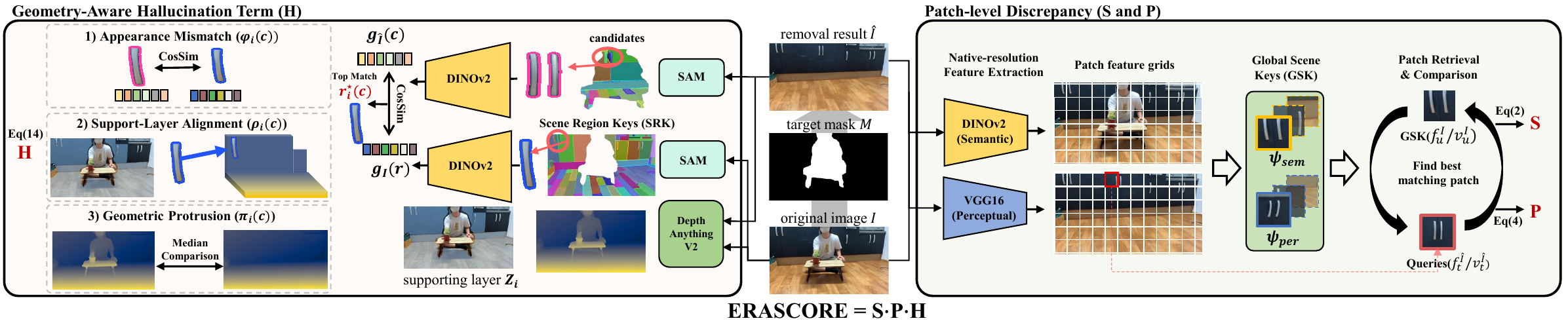}
\caption{\textbf{Overview of ERASCORE.} Patch-level Global Scene Keys (GSK) and region-level Scene Region Keys (SRK) are extracted from the unmasked scene of the original image. The semantic term $S$ and the perceptual term $P$ score patches of the removed region against GSK, and the hallucination term $H$ scores object-like candidate regions against SRK using appearance and depth. The final score is $S\cdot P\cdot H$, and lower is better.}
\label{fig:overview}
\end{figure*}

\subsection{Overview}
\label{sec:overview}

ERASCORE judges a removal without a ground-truth image by testing whether the content a removal method generates in place of the target is supported by the part of the scene that remained visible. It takes an original image $I$ over the pixel domain $\Omega$, a binary target mask $M\subset\Omega$ with connected components $M_1,\ldots,M_N$, and a removal result $\hat I$. We call the part of $\hat I$ inside $M$ the \emph{removed region}, the content that the method places there the \emph{fill}, and the part of $I$ outside $M$ the \emph{unmasked scene}. Fig.~\ref{fig:overview} gives an overview of the metric.

Following the three principles of Sec.~\ref{sec:intro}, the reference is computed from the original image $I$, never from $\hat I$, covers the entire unmasked scene, and keeps scene elements separate, at two granularities: patch-level Global Scene Keys (GSK) and region-level Scene Region Keys (SRK).

Three complementary terms evaluate this evidence. The semantic discrepancy $S$ measures semantic support in DINOv2 space~\cite{oquab2024dinov2}, the perceptual discrepancy $P$ measures rendering mismatch in LPIPS space~\cite{zhang2018unreasonable}, and the geometry-aware hallucination term $H$ examines object-like regions inside the mask. The score is
\begin{equation}
\label{eq:erascore}
\mathrm{ERASCORE}=S\cdot P\cdot H,
\end{equation}
where each term grows with the severity of its failure, so a lower score means a better removal. The term $H$ equals 1 when no region-level penalty applies.

\subsection{Global Scene Keys}
\label{sec:gsk}

Global Scene Keys keep every unmasked patch of the original image as a separate reference, so that a patch of the removed region is supported by whichever scene element it resembles. For a patch encoder $\psi$, each feature-grid position covers a patch of image pixels, its footprint. Let $\mathcal{U}^{\psi}$ be the positions whose footprints lie entirely outside $M$ and $\mathcal{T}^{\psi}$ those whose footprints contain at least one masked pixel, so the two sets partition the grid. Descriptors of $I$ at $\mathcal{U}^{\psi}$ serve as keys and descriptors of $\hat I$ at $\mathcal{T}^{\psi}$ as queries, with $\psi=\mathrm{sem}$ for DINOv2 and $\psi=\mathrm{per}$ for LPIPS. We call the patches at $\mathcal{T}^{\psi}$ \emph{query patches}.

To keep small targets visible, we extract DINOv2 features at native resolution, resizing each image only to the nearest multiple of the patch size. Since a position enters $\mathcal{T}^{\mathrm{sem}}$ as soon as its footprint contains one masked pixel, every mask keeps at least one query, whereas a grid tied to a fixed low resolution can lose a small mask entirely.

\subsection{Semantic Discrepancy (\texorpdfstring{$S$}{S})}
\label{sec:semantic}
$S$ measures the semantic discrepancy between each query patch and its most similar counterpart in the unmasked scene, regardless of where that counterpart lies. To make descriptors reflect coherent regions instead of isolated pixels, we use dense patch tokens from a frozen DINOv2-Giant encoder~\cite{oquab2024dinov2}, whose self-attention incorporates context into each token. Let $f_q^{X}$ be the $\ell_2$-normalized DINOv2 descriptor of image $X\in\{I,\hat I\}$ at grid position $q$. Queries are $f_t^{\hat I}$ for $t\in\mathcal{T}^{\mathrm{sem}}$, and keys are $f_u^{I}$ for $u\in\mathcal{U}^{\mathrm{sem}}$. The support $s_t$ of a query is its highest cosine similarity to any key, clamped at zero, and $S$ is the mean shortfall from full support:
\begin{equation}
\label{eq:semantic}
\begin{aligned}
s_t&=\max_{u\in\mathcal{U}^{\mathrm{sem}}}\max\bigl(0,\cos(f_t^{\hat I},f_u^{I})\bigr),\\
S&=1-\frac{1}{|\mathcal{T}^{\mathrm{sem}}|}\sum_{t\in\mathcal{T}^{\mathrm{sem}}}s_t .
\end{aligned}
\end{equation}
Since $s_t\in[0,1]$, $S$ also lies in $[0,1]$.
Taking the maximum treats GSK as a memory bank, as in patch-based anomaly detection~\cite{roth2022towards}. One matching scene patch is enough to support a query patch, wherever it lies, so a rare but genuine surface is not penalized for being rare, while content that belongs to nothing in the scene finds no good match. The average runs over $\mathcal{T}^{\mathrm{sem}}$ alone, so the unchanged background cannot dilute a small violation inside the mask.

\subsection{Perceptual Discrepancy (\texorpdfstring{$P$}{P})}
\label{sec:perceptual}

$P$ tests how the removed region is rendered, a property that semantic features tend to ignore. DINOv2 inherits the view augmentations of DINO~\cite{caron2021emerging} and iBOT~\cite{zhou2022ibot}, which include Gaussian blur, and it reports strong results on ImageNet-C, whose corruptions include blur~\cite{oquab2024dinov2,hendrycks2019benchmarking}. The same robustness lets a smeared or seamed fill keep high semantic support, and PROVE observes that ReMOVE and CFD likewise score a fill as better as it becomes more blurred~\cite{li2026prove}. To expose such fills, we repeat the retrieval in a space calibrated to human judgments of low-level similarity, LPIPS with a VGG16 backbone~\cite{zhang2018unreasonable,simonyan2014very}. Let $F_l^{X}(z)$ be the channel vector of the $l$-th LPIPS layer at location $z$ for image $X$. Each vector is channel-normalized and reweighted elementwise by $\sqrt{w_l}$, the square root of the nonnegative per-channel LPIPS weight of layer $l$,
\begin{equation}
\label{eq:lpips_feat}
\widehat F_l^{X}(z)=\sqrt{w_l}\odot\frac{F_l^{X}(z)}{\lVert F_l^{X}(z)\rVert_2+\epsilon},
\end{equation}
where $\epsilon=10^{-8}$ guards against division by zero here and in the equations below. With this weighting, the squared distance between two such vectors equals the LPIPS distance at that layer and location.
Pooling $\widehat F_l^{X}$ to a common $64\times64$ grid and concatenating across layers gives a perceptual descriptor $v_q^{X}$ at each grid position $q$, and $\mathcal{T}^{\mathrm{per}}$ and $\mathcal{U}^{\mathrm{per}}$ are defined on this grid as in Sec.~\ref{sec:gsk}. $P$ is the mean squared distance from each query to its nearest key:
\begin{equation}
\label{eq:perceptual}
P=\frac{1}{|\mathcal{T}^{\mathrm{per}}|}\sum_{t\in\mathcal{T}^{\mathrm{per}}}\min_{u\in\mathcal{U}^{\mathrm{per}}}\bigl\lVert v_t^{\hat I}-v_u^{I}\bigr\rVert_2^2 .
\end{equation}
Unlike $S$, $P$ compares descriptors by squared distance and keeps their magnitude, following the LPIPS distance on which it is built.

\subsection{Geometry-Aware Hallucination Term (\texorpdfstring{$H$}{H})}
\label{sec:hallucination}

$H$ penalizes hallucinated structures that patch-level tests miss, by reading candidate regions inside the mask as wholes and checking both what they look like and where they sit in depth. Patch-level retrieval has a blind spot here, and it matters most for strong removal models, whose hallucinations can resemble content already in the scene (Sec.~\ref{sec:analysis_h}). When a method fills the mask with a copy of an object that exists elsewhere in the scene, the generated patches retrieve the real object with high similarity, and the copy is well rendered, as for CLIPAway in Fig.~\ref{fig:qualitative}. The error becomes more apparent at the region level, where the patches form a coherent object that stands in front of the surface that should continue behind the removed target. Unlike CFD~\cite{yu2025omnipaint}, which compares proposals only with adjacent regions of the result, $H$ compares them with the original scene and measures their depth.

\paragraph{Candidates.}
$H$ examines only the object-like regions that the result forms inside the mask. We run the SAM automatic mask generator~\cite{kirillov2023segany} on $\hat I$, denote its proposals by $\mathcal{O}^{\hat I}$, and take the proposals that lie mostly inside component $M_i$ as its candidates,
\begin{equation}
\label{eq:candidates}
\mathcal{C}_i=\Bigl\{c\in\mathcal{O}^{\hat I}:\tfrac{|c\cap M_i|}{|c|}\ge\tau_{\mathrm{cand}}\Bigr\}.
\end{equation}
We set $\tau_{\mathrm{cand}}=0.9$ instead of requiring full containment, because mask boundaries are imprecise and generated content can spill slightly past them. A textureless fill may yield no qualifying proposal, in which case it is left to $S$ and $P$.

\paragraph{Scene Region Keys.}
A candidate is not automatically a hallucination, since a correctly restored background can also form a coherent SAM proposal. To decide whether a candidate belongs to the scene, we compare it with regions of the original image. SAM proposals $\mathcal{O}^{I}$ on $I$ are trimmed by the target mask and by the candidate footprints,
\begin{equation}
\label{eq:srk}
\begin{aligned}
E_i&=M\cup\bigcup_{c\in\mathcal{C}_i}c,\\
\mathcal{R}_i&=\bigl\{m\setminus E_i \;\big|\; m\in\mathcal{O}^{I},\ m\setminus E_i\neq\emptyset\bigr\}.
\end{aligned}
\end{equation}
The trimming keeps the removed target, or a candidate that extends past the mask boundary, from directly serving as evidence for its own reappearance. For a pixel region $r$, let $\mathcal{G}(r)$ be the DINOv2 grid positions whose footprints overlap $r$, and define its descriptor as
\begin{equation}
\label{eq:region_desc}
g_X(r)=\frac{\sum_{q\in\mathcal{G}(r)}f_q^{X}}{\bigl\lVert\sum_{q\in\mathcal{G}(r)}f_q^{X}\bigr\rVert_2+\epsilon},\qquad X\in\{I,\hat I\}.
\end{equation}
Aggregation is appropriate here because the unit under evaluation is a region, and the regions themselves remain separate references. Because DINOv2 tokens are contextualized, the same pixels yield different region descriptors in different surroundings. A pixel-exact copy of an object therefore need not retrieve its source (Sec.~\ref{sec:analysis_h}). A candidate $c$ is described in the result, $g_{\hat I}(c)$, and scene regions in the original, $g_I(r)$. Each candidate retrieves its closest scene region $r^\star_i(c)=\argmax_{r\in\mathcal{R}_i}\cos\bigl(g_{\hat I}(c),g_I(r)\bigr)$. If $\mathcal{R}_i=\emptyset$, which happens only when the mask and its candidates cover every region SAM finds in $I$, we set $\varphi_i(c)=1$ and $\rho_i(c)=0$ below.

\paragraph{Appearance mismatch.}
The base penalty measures how foreign a candidate is to the scene,
\begin{equation}
\label{eq:phi}
\varphi_i(c)=\clip_{[0,1]}\Bigl(1-\cos\bigl(g_{\hat I}(c),g_I(r^\star_i(c))\bigr)\Bigr).
\end{equation}
Appearance alone cannot tell whether a matched structure belongs at the target location. The copied balls in Fig.~\ref{fig:qualitative} resemble the real balls closely ($\varphi=0.23$), and a correctly restored but unusual surface may resemble nothing at all. We therefore add two geometric signals that measure where the matched support lies and where the candidate itself sits.

\paragraph{Supporting layer.}
The supporting layer $Z_i$, which we adopt from the Z-layer of ZCFD~\cite{lee2026zcfd}, tells $H$ where the background should continue behind the target. Let $d_I$ and $d_{\hat I}$ be relative inverse-depth maps from Depth Anything V2~\cite{depth_anything_v2}, where larger values are closer to the camera. For a pixel set $A$, $d_X(A)$ denotes the values of $d_X$ over $A$, and $Q_k(\cdot)$ their $k$-th percentile. A proposal $m\in\mathcal{O}^{I}$ joins $Z_i$ when it touches $M_i$, does not lie mostly inside it, and does not stand clearly in front of the target,
\begin{equation}
\label{eq:zlayer}
\begin{gathered}
m\cap M_i\neq\emptyset,\qquad
\tfrac{|m\cap M_i|}{|m|}<\tau_{\mathrm{ov}},\\
Q_{10}\bigl(d_I(m)\bigr)\le Q_{90}\bigl(d_I(M_i)\bigr),
\end{gathered}
\end{equation}
and $Z_i$ is the union of $m\setminus M_i$ over the admitted proposals, with $\tau_{\mathrm{ov}}=0.8$. The second condition discards fragments of the target's own segmentation. The third rejects $m$ only when at least 90\% of its pixels lie in front of the nearest tenth of the target. When no proposal qualifies, $Z_i$ falls back to the pixels outside $M_i$ with $d_I\le Q_{90}(d_I(M_i))$ that are 8-connected to it, and then to the one-pixel ring around $M_i$, so it is never empty. ZCFD uses this layer as a hard restriction on where appearance may be compared. ERASCORE retrieves appearance from the whole scene and uses $Z_i$ only as a soft prior, because the most similar evidence for a valid fill can lie off that surface (Fig.~\ref{fig:teaser}a).

\paragraph{Support-layer alignment.}
To avoid penalizing valid background that happens to form a proposal, we relieve the penalty when the matched support lies on the supporting layer,
\begin{equation}
\label{eq:rho}
\rho_i(c)=\frac{|r^\star_i(c)\cap Z_i|}{|r^\star_i(c)|}.
\end{equation}
A candidate whose closest match lies on $Z_i$ resembles material at the place where background is expected.

\paragraph{Geometric protrusion.}
To amplify the penalty for content that stands in front of the expected surface, we compare the depth of the candidate with that of the supporting layer. The two depth maps are estimated independently, so each is first standardized by the median and interquartile range of its values over the unmasked region $\Omega\setminus M$,
\begin{equation}
\label{eq:depth_norm}
\widetilde d_X(x)=\frac{d_X(x)-\med\bigl(d_X(\Omega\setminus M)\bigr)}{\IQR\bigl(d_X(\Omega\setminus M)\bigr)+\epsilon},\qquad x\in\Omega,
\end{equation}
and protrusion is the margin by which the candidate stands closer,
\begin{equation}
\label{eq:pi}
\begin{aligned}
\pi_i(c)=\max\Bigl(0,\ &\med\bigl(\widetilde d_{\hat I}(c\cap M_i)\bigr)\\
&-\med\bigl(\widetilde d_{I}(Z_i)\bigr)\Bigr).
\end{aligned}
\end{equation}
Medians keep a few noisy boundary pixels from deciding the order of two surfaces. The copied balls in Fig.~\ref{fig:qualitative} stand about seven interquartile ranges in front of the table ($\pi=7.2$), which turns their small appearance mismatch into a substantial penalty.

\paragraph{Aggregation.}
The three signals form a per-candidate penalty,
\begin{equation}
\label{eq:eta}
\eta_i(c)=\varphi_i(c)\bigl(1-\lambda_{\mathrm{sup}}\rho_i(c)\bigr)\bigl(1+\lambda_{\mathrm{pro}}\pi_i(c)\bigr),
\end{equation}
with $\lambda_{\mathrm{sup}}=0.5$ and $\lambda_{\mathrm{pro}}=1.0$. Alignment can relieve at most half of the appearance mismatch and never cancels it, and protrusion enters at unit scale in the standardized depth units of Eq.~\eqref{eq:depth_norm}. In the CLIPAway result of Fig.~\ref{fig:qualitative}, a small candidate covering tabletop between the copied balls has $\rho=\V{0.98}$ and $\pi=0$, so alignment halves its appearance mismatch of $\varphi=\V{0.35}$ to a penalty of $\eta=\V{0.18}$, whereas the four ball candidates receive \V{1.8} to \V{2.0}. These are values of single candidates, whereas Fig.~\ref{fig:qualitative} reports area-weighted values per output. Geometry thus scales the appearance penalty but never replaces it, so a candidate that matches its surroundings perfectly ($\varphi=0$) receives no penalty however it sits in depth, a case that is rare in the data we evaluate (Sec.~\ref{sec:analysis_h}). Candidate penalties are weighted by their overlap with the component, and components by their area,
\begin{equation}
\label{eq:H}
\begin{aligned}
H_i&=\clip_{[0,1]}\Bigl(\frac{1}{|M_i|}\sum_{c\in\mathcal{C}_i}|c\cap M_i|\,\eta_i(c)\Bigr),\\
H&=1+\frac{\sum_i|M_i|H_i}{\sum_i|M_i|},
\end{aligned}
\end{equation}
with $H_i=0$ when $\mathcal{C}_i$ is empty. SAM proposals may overlap, so $H_i$ is clipped to keep overlapping candidates from inflating the penalty without limit. As a result $H$ lies in $[1,2]$ and acts as a bounded modifier of the patch-level score. Once a component saturates, further ordering is left to $S$ and $P$.

\subsection{Score Combination}
\label{sec:combination}

We multiply the three terms so that they combine without weights fitted to human rankings and without statistics of the evaluated set. Since $S$ and $P$ are positive on every result we evaluate, in log space
\begin{equation}
\label{eq:logsum}
\log\mathrm{ERASCORE}=\log S+\log P+\log H,
\end{equation}
so rescaling any term by a positive constant shifts every result of a scene by the same amount and leaves the ranking within the scene unchanged. A weighted sum requires weights to reconcile the units of $S$ and $P$. The product also fixes the role of $H$, since $H\ge1$ means that region-level evidence can preserve or raise the patch-level penalty but never lower it. We do not claim that the product is a strict logical conjunction, since a near-zero term could in principle hide the others. Sec.~\ref{sec:analysis_combination} compares the product with alternative fusion rules.

\subsection{ERASCORE-T: A Lightweight Variant}
\label{sec:erascore_t}

ERASCORE-T replaces $H$ with a Target Score $T$ that needs neither segmentation nor depth, the two models that dominate the cost of the metric (Sec.~\ref{sec:analysis_t}). $T$ reuses the DINOv2 features that $S$ already computes, adds no threshold or coefficient of its own, and tests a narrower property than $H$, namely whether information about the original target survives in the result. For each query $t\in\mathcal{T}^{\mathrm{sem}}$, we find the nearest background key with the same retrieval as Eq.~\eqref{eq:semantic},
\begin{equation}
\label{eq:t_match}
u^\star(t)=\argmax_{u\in\mathcal{U}^{\mathrm{sem}}}\cos\bigl(f_t^{\hat I},f_u^{I}\bigr),
\end{equation}
and measure two deviations from that background, the footprint of the target and the residual left after removal,
\begin{equation}
\label{eq:t_dev}
\delta^{\mathrm{tgt}}(t)=\bar f_t^{I}-\bar f_{u^\star(t)}^{I},\qquad
\delta^{\mathrm{res}}(t)=\bar f_t^{\hat I}-\bar f_{u^\star(t)}^{I},
\end{equation}
where $\bar f_q^{X}$ denotes the unnormalized DINOv2 token of $X$ at position $q$, so that the deviations keep their magnitude. Here $\bar f_t^{I}$ reads the original image inside the mask, where the target is still visible. The per-patch leakage $\gamma(t)$ combines the direction and the size of the residual relative to the footprint,
\begin{equation}
\label{eq:t_leak}
\begin{aligned}
\gamma(t)={}&\ReLU\Bigl(\cos\bigl(\delta^{\mathrm{res}}(t),\delta^{\mathrm{tgt}}(t)\bigr)\Bigr)\\
&\cdot\frac{\lVert\delta^{\mathrm{res}}(t)\rVert_2}{\lVert\delta^{\mathrm{tgt}}(t)\rVert_2+\epsilon},
\end{aligned}
\end{equation}
and the score mirrors Eq.~\eqref{eq:erascore},
\begin{equation}
\label{eq:t_score}
\begin{gathered}
T=1+\frac{1}{|\mathcal{T}^{\mathrm{sem}}|}\sum_{t\in\mathcal{T}^{\mathrm{sem}}}\gamma(t),\\
\mathrm{ERASCORE}\text{-}\mathrm{T}=S\cdot P\cdot T .
\end{gathered}
\end{equation}
The ratio measures how much of the target's departure from its matched background survives, normalized by how distinctive the target was in the first place. The ReLU gate restricts the penalty to residuals that point toward the target itself. Without it, a negative cosine would let unrelated content pull $T$ below one and lower the patch-level penalty, and residual and newly generated patches could cancel within the same mask. The gate keeps $T\ge1$, the same role that $H$ plays in Eq.~\eqref{eq:erascore}, but $T$ looks for what remains of the target, whereas $H$ looks for structures that appear in its place.

\section{Experimental Setup}
\label{sec:setup}

\paragraph{Benchmarks and removal methods.}
We evaluate on \nCases{} samples, each an image and a target mask, from RORD-Val (343), OBER-Test (163), and OBER-Wild (302, without ground truth). RORD-Val is the split of RORD~\cite{sagong2022rord} that ObjectClear~\cite{zhao2026objectclear} builds by keeping one image per scene and refining its masks, and OBER-Test and OBER-Wild are the test sets of OBER~\cite{zhao2026objectclear}. We refer to RORD-Val as RORD in the tables. To cover a wide range of output quality, each sample is processed by eight methods that span both task specialization and recency. Six are built for object removal, from an established representative, CLIPAway~\cite{ekin2024clipaway}, to current systems, OmniEraser~\cite{Wei2025OmniEraserRO}, ObjectClear~\cite{zhao2026objectclear}, YOEO~\cite{YOEO}, OmniPaint~\cite{yu2025omnipaint}, and RORem~\cite{Li2025RORemTA}. Two are general inpainting models that can be applied to removal without adaptation, the recent Moebius~\cite{DuanAndXu2026Moebius} and the established BrushNet~\cite{Ju2024BrushNetAP} with an SDXL~\cite{podell2024sdxl} backbone.

\paragraph{Human rankings.}
The human target is a ranking of all eight outputs of a sample, so that raters compare the methods directly on the same scene instead of scoring images in isolation. We recruited \nRatersAll{} participants through Prolific~\cite{prolific}, and each ranked the eight outputs of 15 to 17 samples from best to worst by dragging them into order, viewing the original image and the target mask, with method identities hidden and the starting order shuffled. We prioritized coverage so that every one of the \nCases{} samples would receive a human ranking, and 355 samples, assigned at random, were ranked by four independent participants. These make up \V{42.9\%} to \V{44.9\%} of each benchmark, and the study yielded \nRankingsAll{} complete rankings, one on each of the other 453 samples. We use the mean of the available ranks as the \emph{human reference ranking} of a sample. Single-ranking targets are noisier than averaged ones, but every metric is scored against the same targets, so all metrics face the same noise, and Sec.~\ref{sec:results_mor} shows that the conclusions do not depend on the number of rankings per sample.

\paragraph{Evaluation protocol.}
For each sample, we compute Spearman's (SRCC) and Kendall's $\tau_b$~\cite{kendall1945treatment} (KRCC) rank correlation between a metric's ordering of the eight outputs and the human reference ranking, and we average over samples. We compare two metrics with a two-sided paired Wilcoxon signed-rank test~\cite{wilcoxon1945individual} on unrounded per-sample correlations over the samples on which both are defined, and we apply Holm correction~\cite{holm1979simple} when one claim rests on several tests. The 355 samples with four rankings also give the inter-rater agreement, the mean SRCC over all pairs of their raters, which is \ceiling{}. This value compares one individual with another, whereas a metric is scored against the mean of the available ranks, so it is not an upper bound on what a metric can reach.

\paragraph{Baselines.}
We compare against \nBaselines{} baselines in four groups. Four are reference-free removal metrics, ZCFD~\cite{lee2026zcfd}, PROVE RC-S~\cite{li2026prove}, ReMOVE~\cite{chandrasekar2024remove}, and CFD~\cite{yu2025omnipaint}. ReMOVE is computed on a square crop around the mask, the variant its authors adopt as ReMOVE, and Table~\ref{tab:main} also lists the variant computed on the full image, which they report as ReMOVE without cropping. Sixteen are full-reference metrics, including DreamSim~\cite{fu2023dreamsim}, LPIPS~\cite{zhang2018unreasonable}, and SSIM~\cite{wang2004image}, evaluated on the 506 samples with ground truth. Seven are no-reference quality metrics, including MANIQA~\cite{yang2022maniqa} and MUSIQ~\cite{ke2021musiq}. The last two are listwise MLLM judges, Gemini-2.5-Pro~\cite{comanici2025gemini} and GPT-4o~\cite{openai2024gpt4o}. Table~\ref{tab:main} lists every baseline with its reference. Each metric is oriented so that a higher correlation means better agreement. The MLLM judges share one prompt, image layout, and random shuffle per sample. They see the original image, an overlay marking the target, and the eight outputs together, record diagnostics for each output (complete removal, hallucinated content, residual shadow or reflection, and artifacts), and then assign a removal-quality score from 1 to 10 together with a full ranking. We report the score field, which is the stronger of the two for both judges. Because these integer scores tie on about \V{80\%} of samples, the KRCC of the judges uses $\tau_b$.

\paragraph{Implementation.}
ERASCORE uses one configuration throughout, with frozen DINOv2-Giant (patch size 14), LPIPS-VGG at an input of 896 pixels pooled to a $64\times64$ grid, SAM ViT-H, and Depth Anything V2 Base. The constants of $H$ are $\tau_{\mathrm{cand}}=0.9$, $\tau_{\mathrm{ov}}=0.8$, $\lambda_{\mathrm{sup}}=0.5$, $\lambda_{\mathrm{pro}}=1.0$, depth percentiles of 90 and 10, and $\epsilon=10^{-8}$. None of them is fitted to human rankings or changes across benchmarks and methods, and agreement is insensitive to them. Perturbing each constant of $H$ up or down moves SRCC by at most \V{0.0013}, and the only significant change raises agreement above the value we use, which is consistent with a configuration that was not tuned for human agreement (Table~\ref{tab:sens}).

\section{Agreement with Human Rankings}
\label{sec:results}

\begin{table*}[t]
\centering
\footnotesize
\caption{\textbf{Agreement with human rankings.} SRCC and KRCC are averaged over samples, Std is the standard deviation of the three benchmark SRCCs, and $n$ is the number of samples on which a metric is defined. The restricted ERASCORE rows use the samples of ZCFD and of the full-reference metrics, which need ground truth (RORD and OBER-Test). ZCFD assigns the same value to all eight outputs on 9 samples, on which the mask or the background region it compares against covers no cell of its 16$\times$16 feature grid. $\dagger$: full-image ReMOVE, undefined on 3 samples with very small masks and not counted among the 29 baselines. The human row is the inter-rater agreement (Sec.~\ref{sec:setup}). Bold marks the best of ERASCORE and the 29 baselines, excluding the other ERASCORE rows. All 29 baselines are significantly below ERASCORE (maximum Holm-adjusted $p=1.56\times10^{-4}$).}
\label{tab:main}
\setlength{\tabcolsep}{5pt}
\begin{tabular}{llrrrrrrr}
\toprule
Type & Metric & RORD & OBER-Test & OBER-Wild & SRCC & Std & KRCC & $n$ \\
\midrule
Ours & \textbf{ERASCORE} & \textbf{0.618} & \textbf{0.705} & \textbf{0.720} & \textbf{0.6739} & 0.045 & \textbf{0.5646} & 808 \\
     & ERASCORE-T & 0.621 & 0.706 & 0.717 & 0.6738 & 0.043 & 0.5649 & 808 \\
     & ERASCORE, ZCFD samples & 0.623 & 0.700 & 0.721 & 0.6751 & 0.042 & 0.5655 & 799 \\
     & ERASCORE, ground-truth samples & 0.618 & 0.705 & -- & 0.6462 & -- & 0.5401 & 506 \\
\midrule
\multirow{2}{*}{MLLM judge} & Gemini-2.5-Pro~\cite{comanici2025gemini} & 0.559 & 0.702 & 0.700 & 0.6405 & 0.067 & 0.5365 & 808 \\
 & GPT-4o~\cite{openai2024gpt4o} & 0.563 & 0.671 & 0.685 & 0.6302 & 0.055 & 0.5274 & 808 \\
\midrule
\multirow{5}{*}{Reference-free} & ZCFD~\cite{lee2026zcfd} & 0.441 & 0.608 & 0.621 & 0.5415 & 0.082 & 0.4378 & 799 \\
 & PROVE RC-S~\cite{li2026prove} & 0.420 & 0.620 & 0.586 & 0.5222 & 0.087 & 0.4120 & 808 \\
 & ReMOVE (crop)~\cite{chandrasekar2024remove} & 0.295 & 0.500 & 0.493 & 0.4103 & 0.095 & 0.3211 & 808 \\
 & CFD~\cite{yu2025omnipaint} & 0.250 & 0.304 & 0.407 & 0.3195 & 0.065 & 0.2475 & 808 \\
 & ReMOVE (full image)$^\dagger$ & 0.098 & 0.492 & 0.541 & 0.3428 & 0.198 & 0.2687 & 805 \\
\midrule
\multirow{16}{*}{Full-reference} & DreamSim~\cite{fu2023dreamsim} & 0.483 & 0.649 & -- & 0.5364 & -- & 0.4394 & 506 \\
 & TOPIQ-FR~\cite{chen2024topiq} & 0.424 & 0.632 & -- & 0.4912 & -- & 0.3930 & 506 \\
 & PSNR & 0.322 & 0.542 & -- & 0.3926 & -- & 0.3100 & 506 \\
 & AHIQ~\cite{lao2022attentions} & 0.311 & 0.536 & -- & 0.3836 & -- & 0.3015 & 506 \\
 & VSI~\cite{zhang2014vsi} & 0.327 & 0.487 & -- & 0.3785 & -- & 0.2950 & 506 \\
 & MS-SSIM~\cite{wang2003multiscale} & 0.303 & 0.483 & -- & 0.3606 & -- & 0.2816 & 506 \\
 & PieAPP~\cite{prashnani2018pieapp} & 0.364 & 0.335 & -- & 0.3548 & -- & 0.2761 & 506 \\
 & GMSD~\cite{xue2014gradient} & 0.297 & 0.459 & -- & 0.3494 & -- & 0.2708 & 506 \\
 & DISTS~\cite{ding2020image} & 0.301 & 0.449 & -- & 0.3489 & -- & 0.2751 & 506 \\
 & NLPD~\cite{laparra2016perceptual} & 0.275 & 0.449 & -- & 0.3307 & -- & 0.2564 & 506 \\
 & LPIPS (Alex)~\cite{zhang2018unreasonable} & 0.271 & 0.409 & -- & 0.3154 & -- & 0.2450 & 506 \\
 & FSIM~\cite{zhang2011fsim} & 0.252 & 0.393 & -- & 0.2975 & -- & 0.2256 & 506 \\
 & MAD~\cite{larson2010most} & 0.158 & 0.266 & -- & 0.1927 & -- & 0.1345 & 506 \\
 & SSIM~\cite{wang2004image} & 0.135 & 0.210 & -- & 0.1594 & -- & 0.1140 & 506 \\
 & LPIPS (VGG)~\cite{zhang2018unreasonable} & 0.111 & 0.171 & -- & 0.1303 & -- & 0.0873 & 506 \\
 & VIF~\cite{sheikh2006image} & 0.028 & 0.026 & -- & 0.0273 & -- & 0.0030 & 506 \\
\midrule
\multirow{7}{*}{No-reference} & MANIQA~\cite{yang2022maniqa} & 0.335 & 0.247 & 0.273 & 0.2942 & 0.037 & 0.2276 & 808 \\
 & TOPIQ-NR~\cite{chen2024topiq} & 0.303 & 0.174 & 0.135 & 0.2140 & 0.072 & 0.1627 & 808 \\
 & MUSIQ~\cite{ke2021musiq} & 0.310 & 0.233 & $-$0.046 & 0.1617 & 0.153 & 0.1261 & 808 \\
 & BRISQUE~\cite{mittal2012no} & 0.139 & 0.054 & 0.041 & 0.0853 & 0.043 & 0.0587 & 808 \\
 & CLIP-IQA~\cite{wang2023exploring} & 0.293 & $-$0.068 & $-$0.191 & 0.0392 & 0.205 & 0.0290 & 808 \\
 & NIQE~\cite{mittal2012making} & 0.034 & 0.060 & $-$0.005 & 0.0245 & 0.026 & 0.0163 & 808 \\
 & NIMA~\cite{talebi2018nima} & 0.150 & $-$0.043 & $-$0.172 & $-$0.0093 & 0.133 & $-$0.0113 & 808 \\
\midrule
Human & Inter-rater agreement & 0.503 & 0.640 & 0.614 & 0.5710 & 0.059 & 0.4781 & 355 \\
\bottomrule
\end{tabular}
\end{table*}

\subsection{Comparison with Baselines}
\label{sec:results_main}

ERASCORE, which draws its evidence from the observed scene, agrees with human judgments better than every alternative in our comparison, including metrics that use a ground-truth image and commercial multimodal judges. It reaches \srccMain{} SRCC over the \nCases{} samples, leads every baseline on each benchmark, and is significantly above all \nBaselines{} baselines (Table~\ref{tab:main}). No-reference metrics, which score naturalness, all fall far below ERASCORE. Reference-free metrics, which read their evidence from the output, a local window, or a pooled summary, do better, and the strongest of them, ZCFD, reaches \srccZCFD{} on the 799 samples on which it is defined, where ERASCORE reaches \V{0.6751}.

A ground-truth image does not close this gap. On the 506 samples that have one, ERASCORE agrees with human rankings more closely than DreamSim (\srccOursGT{} against \srccDreamSim{}, Table~\ref{tab:main}) without ever seeing the ground truth. Distance to a single captured completion can penalize fills that differ from it yet continue the scene correctly, whereas support retrieved from the scene can accept them. The MLLM judges are the strongest baselines, and ERASCORE still exceeds both (\srccGemini{} for Gemini-2.5-Pro), while running locally without external API calls and producing continuous scores.

ERASCORE also varies less across benchmarks than every reference-free baseline (Std in Table~\ref{tab:main}). No coefficient in Eq.~\eqref{eq:erascore} can adapt to a benchmark, so this consistency does not come from per-benchmark tuning.

\begin{table}[t]
\centering
\footnotesize
\caption{\textbf{Agreement with individual raters and boundary accuracy.} Held-out SRCC is the agreement with one held-out rater on the 355 samples with four rankings (Sec.~\ref{sec:results_human}), and $\ast$ marks a metric significantly below the three-rater mean. Against one other rater, ERASCORE ($p=1.33\times10^{-11}$) and both MLLM judges are significantly higher, 26 of the 29 baselines are significantly lower, and DreamSim does not differ. Boundary accuracy (Sec.~\ref{sec:results_boundary}) uses 724, 720, and 768 samples for $k=2,3,4$, and there $\ast$ marks a baseline significantly below ERASCORE after Holm correction over the 29 baselines. The 23 omitted baselines are also significantly below the three-rater mean and below ERASCORE in boundary accuracy. Bold marks the best of ERASCORE and the 29 baselines.}
\label{tab:extra}
\setlength{\tabcolsep}{4pt}
\begin{tabular}{lcccc}
\toprule
 & Held-out & \multicolumn{3}{c}{Boundary accuracy} \\
\cmidrule(lr){3-5}
 & SRCC & $k=2$ & $k=3$ & $k=4$ \\
\midrule
\multicolumn{5}{l}{\textit{Human}} \\
\quad One other rater & 0.571 & -- & -- & -- \\
\quad Three-rater mean & 0.644 & -- & -- & -- \\
\quad Same worst / best & \multicolumn{4}{l}{60.4\% / 31.4\% of rater pairs} \\
\midrule
\textbf{ERASCORE} & \textbf{0.634} & \textbf{0.915} & \textbf{0.879} & \textbf{0.830} \\
ERASCORE-T & 0.634 & 0.915 & 0.879 & 0.831 \\
Gemini-2.5-Pro & 0.614$^\ast$ & 0.903$^\ast$ & 0.863$^\ast$ & 0.812$^\ast$ \\
GPT-4o & 0.600$^\ast$ & 0.895$^\ast$ & 0.859$^\ast$ & 0.808$^\ast$ \\
ZCFD & 0.512$^\ast$ & 0.864$^\ast$ & 0.810$^\ast$ & 0.769$^\ast$ \\
PROVE RC-S & 0.500$^\ast$ & 0.858$^\ast$ & 0.814$^\ast$ & 0.768$^\ast$ \\
ReMOVE & 0.398$^\ast$ & 0.792$^\ast$ & 0.742$^\ast$ & 0.702$^\ast$ \\
CFD & 0.322$^\ast$ & 0.730$^\ast$ & 0.683$^\ast$ & 0.659$^\ast$ \\
\bottomrule
\end{tabular}
\end{table}

\subsection{Agreement Relative to Human Raters}
\label{sec:results_human}

ERASCORE agrees with an individual rater better than another rater does. The inter-rater value compares one rater with another, whereas a metric is compared with the mean of the raters (Sec.~\ref{sec:setup}). To put the two on the same footing, we hold out one rater at a time on the 355 samples with four rankings and ask a metric, a single other rater, and the mean of the other three raters to recover the held-out ranking. ERASCORE predicts it significantly better than a single other rater does, and its agreement is not significantly different from that of the three-rater mean (Table~\ref{tab:extra}).

ERASCORE and ERASCORE-T are also the only metrics that are not significantly worse than the three-rater mean. Most baselines fall below even a single other rater, and the two MLLM judges, which exceed a single rater, still fall significantly below the three-rater mean (Table~\ref{tab:extra}). Not detecting a difference is not the same as showing equality, so we state our claim at the level of an individual rater and not of a human consensus.

\subsection{Number of Rankings per Sample}
\label{sec:results_mor}

\begin{table}[t]
\centering
\footnotesize
\caption{\textbf{Agreement for different numbers of rankings per sample} (SRCC). \emph{All} is the reference used throughout the paper, and \emph{Four rankings} restricts it to the 355 samples that have four. $\dagger$ marks the baseline not significantly below ERASCORE after Holm correction over the 29 baselines. The rows with one to three rankings average that many of the four rankings of each sample, over all subsets of that size. The single-ranking rows replace the four rankings of each of the 355 samples with those of one rater, for each rater in turn. They report Spearman's correlation between the orders of all 30 metrics with and without this replacement, averaged over the four raters, and the number of baselines significantly below ERASCORE in all four cases. ZCFD is evaluated on its 799 samples, 349 of them among the 355.}
\label{tab:mor}
\setlength{\tabcolsep}{4pt}
\begin{tabular}{lcc}
\toprule
Metric & All (808) & Four rankings (355) \\
\midrule
\textbf{ERASCORE} & \textbf{0.6739} & \textbf{0.7432} \\
Gemini-2.5-Pro & 0.6405 & 0.7168$^\dagger$ \\
GPT-4o & 0.6302 & 0.6953 \\
ZCFD & 0.5415 & 0.5961 \\
PROVE RC-S & 0.5222 & 0.5743 \\
\midrule
\multicolumn{3}{l}{\textit{ERASCORE against fewer rankings}} \\
\quad 1 ranking & -- & 0.634 \\
\quad 2 rankings & -- & 0.681 \\
\quad 3 rankings & -- & 0.722 \\
\midrule
\multicolumn{3}{l}{\textit{Single-ranking reference on the 355 samples}} \\
\quad Metric order (Spearman) & 0.9987 & 0.9951 \\
\quad Baselines below ERASCORE & 29 of 29 & 27 of 29 \\
\midrule
Kendall's $W$ & -- & 0.68 \\
\bottomrule
\end{tabular}
\end{table}

ERASCORE agrees more closely with human judgments as more independent rankings are averaged into the reference. On the same 355 samples, its SRCC rises monotonically from 0.634 with one ranking to \srccMORfour{} with four (Table~\ref{tab:mor}). Since averaging reduces the influence of individual differences in judgment, this trend suggests that ERASCORE captures judgments shared across raters. The four rankings of each sample also show agreement with one another, with a mean Kendall's $W$~\cite{kendall1939problem} of 0.68 (Table~\ref{tab:mor}).

The advantage of ERASCORE over the baselines also persists across the reference configurations we evaluate. When the four rankings of each of the 355 samples are replaced by a single ranking, the order of all 30 metrics across the full \nCases{} samples barely changes, and ERASCORE remains first and significantly above all \nBaselines{} baselines for each of the four raters (Table~\ref{tab:mor}). On the subset of 355 samples evaluated against the mean of four rankings, ERASCORE again ranks first and significantly outperforms every baseline except Gemini-2.5-Pro. The five metrics in Table~\ref{tab:mor} retain the same order on the full set and this subset.

\begin{figure*}[t]
\centering
\includegraphics[width=\linewidth]{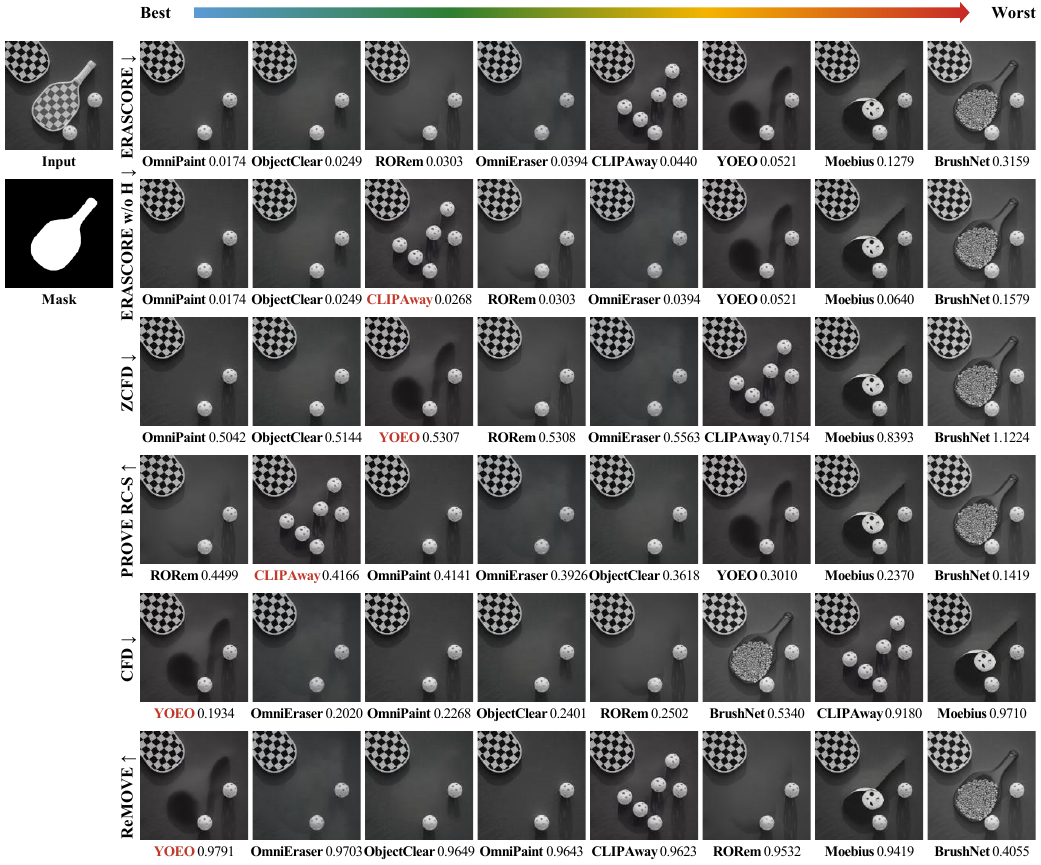}
\vspace{3pt}\par
{\footnotesize
\setlength{\tabcolsep}{0pt}
\renewcommand{\arraystretch}{1.1}
\begin{tabular}{
  >{\raggedright\arraybackslash}p{0.13786\linewidth}
  *{7}{>{\centering\arraybackslash}p{0.10714\linewidth}}
  >{\centering\arraybackslash}p{0.11214\linewidth}}
\toprule
\hspace{0.00929\linewidth}visible
  & clean & brighter, softer & texture lost, warm & color shift
  & balls copied & shadow painted & malformed ball & residue + copy \\
\midrule
\hspace{0.00929\linewidth}$S$
  & 0.32 & 0.39 & 0.41 & 0.42 & \textbf{0.31} & 0.50 & 0.54 & 0.67 \\
\hspace{0.00929\linewidth}$P$
  & 0.054 & 0.064 & 0.075 & 0.093 & 0.086 & 0.104 & 0.119 & 0.237 \\
\hspace{0.00929\linewidth}$H$
  & 1.00 & 1.00 & 1.00 & 1.00 & \textbf{1.64} & 1.00 & \textbf{2.00} & \textbf{2.00} \\
\midrule
\hspace{0.00929\linewidth}$\varphi$
  & -- & -- & -- & -- & 0.23 & -- & 0.64 & 0.67 \\
\hspace{0.00929\linewidth}$\rho$
  & -- & -- & -- & -- & 0.02 & -- & 0.58 & 0.99 \\
\hspace{0.00929\linewidth}$\pi$
  & -- & -- & -- & -- & 7.2 & -- & 3.0 & 4.7 \\
\bottomrule
\end{tabular}}
\caption{\textbf{Removing a pickleball paddle from a tabletop} (OBER-Wild). Each row orders the eight outputs by one reference-free metric, best at left, with the raw value under each image. The table follows the ERASCORE row. The first line describes what is visible in each output, and $\varphi$, $\rho$, and $\pi$ are the area-weighted sub-terms of $H$, shown where a candidate region was found. Red names mark, in each row, an output containing generated content that the metric places among the four cleanest.}
\label{fig:qualitative}
\end{figure*}

\subsection{Boundary Accuracy}
\label{sec:results_boundary}

ERASCORE and its lightweight variant are the most accurate metrics at separating the outputs that humans rank highest from those they agree have failed. This boundary matters because humans agree on what failed far more than on what succeeded best. Two raters of the same sample pick the same worst output in about three fifths of cases, but the same best output in fewer than a third (Table~\ref{tab:extra}). We therefore also count how often a metric orders a human top-$k$ output above a human bottom-$k$ output, excluding samples where tied human ranks straddle the boundary. ERASCORE leads every baseline for $k=2,3,4$, and its margin over the strongest baseline, Gemini-2.5-Pro, is small but significant for every $k$ (Table~\ref{tab:extra}).

\section{Qualitative Analysis}
\label{sec:qualitative}

Fig.~\ref{fig:qualitative} shows how the three terms divide the work on one sample. All four raters place the same four outputs in the upper half, although they differ on which output is best and which is worst.

\paragraph{Clean fills.}
$S$ and $P$ track visible degradation as long as the fill remains background. The four outputs in the upper half degrade in visible steps. OmniPaint preserves the texture and color of the table, ObjectClear's fill is slightly brighter and softer, RORem's loses the texture, turns warmer, and leaves the occluded edge of a ball unrestored, and OmniEraser shifts the color further. $S$ and $P$ rise in exactly this order with $H=1$ throughout. Of the four raters, one gives exactly this order, one places ObjectClear last of the four, and two place RORem last.

\paragraph{Generated content.}
$H$ corrects the one failure that patch retrieval prefers. CLIPAway fills the mask with close copies of a ball from elsewhere on the table. The semantic term rates this output as the best supported of the eight ($S=0.31$, the lowest), and without $H$ it ranks third. The copies match the real balls ($\varphi=0.23$), but they stand seven interquartile ranges in front of the table ($\pi=7.2$) and find almost no support on the expected surface ($\rho=0.02$). $H$ multiplies the score by 1.64 and moves the output from the clean group to the failed one, raising the per-sample SRCC from 0.738 to 0.905. Moebius inserts a malformed ball and BrushNet a structure that merges residue with copied content. Both show large appearance mismatch and protrusion, both saturate $H$, and $S$ and $P$ then set their order. YOEO paints a shadow into the mask as a large textureless region that yields no proposal, so its penalty comes from $S$ and $P$ alone. The copied balls stop at fifth place, short of the seventh place they receive from the mean of the four raters, because they cover a third of the mask and $H$ is bounded at 2 (Eq.~\ref{eq:H}). The CLIPAway result illustrates the failure mode examined in Sec.~\ref{sec:analysis_h}: a plausible hallucination can receive high patch-level support when it resembles existing scene content. By examining the generated structure as a whole and accounting for its depth relative to the expected background, $H$ penalizes a failure that $S\cdot P$ alone does not sufficiently penalize.

\paragraph{Baseline failures.}
The failures the baselines promote here are consistent with how they build their references. Among the reference-free metrics shown, only ERASCORE places all four outputs with generated content after the four clean fills. ZCFD, CFD, and ReMOVE rank YOEO's painted shadow among the clean outputs, third, first, and first respectively. None of the three separates the dark, textureless fill from the tabletop. PROVE RC-S instead ranks the copied balls second, plausibly because copied scene content matches the feature statistics around the target, which a distribution-level comparison rewards. ZCFD, PROVE RC-S, and ReMOVE do place the malformed and merged structures last, as ERASCORE does.

\section{Ablation Study and Analysis}
\label{sec:analysis}

This section tests the design decisions of Sec.~\ref{sec:method} one at a time, changing a single decision and holding everything else fixed. The reference-design experiments use $S\cdot P$, which isolates the reference from the region branch.

\begin{table}[t]
\centering
\footnotesize
\caption{\textbf{Reference design} ($S\cdot P$, SRCC). Each block changes one property of the reference with the others held at the adopted setting (bold). Pooled rows replace the references of $S$, $P$, or both with a single mean descriptor. $\ast$ marks a row significantly below the adopted setting.}
\label{tab:reference}
\setlength{\tabcolsep}{3.6pt}
\begin{tabular}{lcccc}
\toprule
 & RORD & OBER-Test & OBER-Wild & All \\
\midrule
\multicolumn{5}{l}{\textit{Source}} \\
Removal output $\hat I$ & 0.508 & 0.662 & 0.683 & 0.6045$^\ast$ \\
\textbf{Original image $I$} & \textbf{0.622} & \textbf{0.706} & \textbf{0.711} & \textbf{0.6720} \\
\midrule
\multicolumn{5}{l}{\textit{Scope}} \\
Local crop (ReMOVE) & 0.613 & 0.692 & 0.696 & 0.6601$^\ast$ \\
Mask box (CFD) & 0.621 & 0.681 & 0.706 & 0.6645$^\ast$ \\
\textbf{Full scene} & \textbf{0.622} & \textbf{0.706} & \textbf{0.711} & \textbf{0.6720} \\
\midrule
\multicolumn{5}{l}{\textit{Representation}} \\
Pooled $S$ only & 0.597 & 0.685 & 0.702 & 0.6541$^\ast$ \\
Pooled $P$ only & 0.620 & 0.704 & 0.711 & 0.6709 \\
Pooled both & 0.458 & 0.658 & 0.662 & 0.5749$^\ast$ \\
\textbf{Separate} & \textbf{0.622} & \textbf{0.706} & \textbf{0.711} & \textbf{0.6720} \\
\bottomrule
\end{tabular}
\end{table}

\begin{table}[t]
\centering
\footnotesize
\caption{\textbf{Term contributions} (SRCC, 808 samples). $\Delta$ is the change from the full metric. $H$ alone is defined on 804 samples.}
\label{tab:terms}
\setlength{\tabcolsep}{6pt}
\begin{tabular}{lcc}
\toprule
Configuration & SRCC & $\Delta$ \\
\midrule
\textbf{$S\cdot P\cdot H$ (ERASCORE)} & \textbf{0.6739} & -- \\
$P\cdot H$ (without $S$) & 0.6345 & $-$0.0394 \\
$S\cdot H$ (without $P$) & 0.6594 & $-$0.0145 \\
$S\cdot P$ (without $H$) & 0.6720 & $-$0.0019 \\
\midrule
$S$ alone & 0.6427 & $-$0.0312 \\
$P$ alone & 0.6219 & $-$0.0520 \\
$H$ alone & 0.4975 & $-$0.1764 \\
\bottomrule
\end{tabular}
\end{table}

\subsection{Reference Design}
\label{sec:analysis_reference}

\paragraph{Source.}
Reading the reference from the original image raises agreement on every benchmark. Reading the keys from $\hat I$ instead, as ReMOVE, CFD, and PROVE do for their context, lowers agreement on all three benchmarks and most on RORD (Table~\ref{tab:reference}). With keys from $\hat I$, a fill that smears a boundary or shifts the color of nearby background also changes the evidence it is scored against, and the comparison slides toward a test of self-consistency.

\paragraph{Scope.}
Searching the whole unmasked scene for support beats the local windows of existing metrics, which cannot reach material that lies far from the mask. Restricting the keys to the local crop of ReMOVE or to the box around the mask used in CFD's official implementation lowers agreement significantly in both cases, and the full scene is best on every benchmark. The effect is smaller than that of the source, which suggests that suitable support often lies near the mask, and the full scene matters when it does not, as for the gallery labels in Fig.~\ref{fig:teaser}a.

\paragraph{Representation.}
Keeping scene elements separate matters most of the three choices. Covering the full scene does not help if its elements are merged into one descriptor, and pooling the references of both terms into a single mean causes the largest loss in Table~\ref{tab:reference}. Pooling only one of them costs much less, and pooling only $P$ makes no significant difference, which suggests that the two terms compensate for each other and that the term that still retrieves recovers much of the local evidence. This fallback disappears once both are pooled, in line with the observation of \citet{li2026prove} that averaging can dilute localized structural errors.

\subsection{Semantic and Perceptual Terms}
\label{sec:analysis_terms}

Semantic and perceptual support capture different failures, and each is strong on its own. $S$ alone already reaches the level of the MLLM judges, and $P$ alone exceeds every baseline other than the MLLM judges (Table~\ref{tab:terms}). Removing either from the full metric costs agreement, $S$ most of all, so each carries information the other lacks.

The semantic term does not need the largest encoder. Replacing DINOv2-Giant with DINOv2-Base, at less than a tenth of the parameters, lowers SRCC by only \V{0.002}, a difference a paired test does not detect, and DINOv2-Large falls slightly below both ($p=0.027$ against Giant before correction, Table~\ref{tab:robust}).

\subsection{Score Combination}
\label{sec:analysis_combination}

\begin{table}[t]
\centering
\footnotesize
\caption{\textbf{Fusion rules} (SRCC, 808 samples). The last column marks rules that normalize the terms with statistics of the whole evaluated set. Rules marked with $^\ast$ are significantly below the product.}
\label{tab:fusion}
\setlength{\tabcolsep}{5pt}
\begin{tabular}{lcc}
\toprule
Rule & SRCC & Dataset statistics \\
\midrule
\textbf{Product $S\cdot P\cdot H$ (ours)} & \textbf{0.6739} & -- \\
Mixed $(S+P)\cdot H$ & 0.6731 & -- \\
Sum & 0.6654$^\ast$ & -- \\
Per-sample z-score sum & 0.6638$^\ast$ & -- \\
Per-sample rank mean & 0.6544$^\ast$ & -- \\
Per-sample min-max, maximum & 0.6497$^\ast$ & -- \\
\midrule
Dataset z-score sum & 0.6753 & \checkmark \\
Dataset min-max, mean & 0.6724 & \checkmark \\
Dataset min-max, soft OR & 0.6705 & \checkmark \\
Dataset min-max, maximum & 0.6415$^\ast$ & \checkmark \\
\bottomrule
\end{tabular}
\end{table}

\begin{table}[t]
\centering
\footnotesize
\caption{\textbf{Sensitivity to the constants of $H$ and to log-space weights} (808 samples). (a) Change in SRCC when one constant is moved down or up, with $\ast$ for a significant change. (b) Weights $(w_S,w_P,w_H)$ on $\log S$, $\log P$, and $\log H$ over 91 settings, compared with the product.}
\label{tab:sens}
\setlength{\tabcolsep}{4pt}
\begin{tabular}{lcrr}
\toprule
\multicolumn{4}{l}{\textit{(a) Constants of $H$}} \\
Constant (adopted) & Tested & Down & Up \\
\midrule
$\tau_{\mathrm{cand}}$ (0.90) & 0.85, 0.95 & +0.0013 & +0.0001 \\
$\tau_{\mathrm{ov}}$ (0.80) & 0.70, 0.90 & 0.0000 & $-$0.0001 \\
Target percentile (90) & 85, 95 & 0.0000 & $-$0.0001 \\
Proposal percentile (10) & 5, 15 & $-$0.0002 & $-$0.0002 \\
$\lambda_{\mathrm{sup}}$ (0.5) & 0.3, 0.7 & +0.0006 & +0.0001 \\
$\lambda_{\mathrm{pro}}$ (1.0) & 0.5, 1.5 & 0.0000 & +0.0011$^\ast$ \\
\midrule
\multicolumn{4}{l}{\textit{(b) Log-space weights}} \\
\multicolumn{2}{l}{Best setting $(1, 0.75, 1.25)$} & \multicolumn{2}{r}{+0.0024 ($p=0.049$)} \\
\multicolumn{2}{l}{Settings above the product, uncorrected} & \multicolumn{2}{r}{2 of 91} \\
\multicolumn{2}{l}{Settings above the product, Holm} & \multicolumn{2}{r}{0 of 91} \\
\bottomrule
\end{tabular}
\end{table}

Among the rules that need neither fitted weights nor statistics of the evaluated set, the product agrees best with human rankings, and no rule we tested does significantly better (Table~\ref{tab:fusion}). Rules that put the terms on a common scale for each sample, by z-scores, ranks, or min-max scaling, all fall significantly below it. The rules that come close normalize the terms with statistics of the whole evaluated set, so the score of one output would change whenever other outputs were added to or removed from the evaluation. The mixed rule $(S+P)\cdot H$ avoids such statistics but adds $S$ and $P$ directly, so it depends on their relative scale (Sec.~\ref{sec:combination}). Fitting weights in log space brings no significant gain after Holm correction over the 91 weight settings we tested, and any weights found that way would be fitted to the human rankings themselves (Table~\ref{tab:sens}).

\begin{table}[t]
\centering
\footnotesize
\caption{\textbf{Real pixels pasted into the mask} (808 samples per construction). $\Delta\log$ is the change in log penalty relative to the mean over the eight real removals, so a positive value means the term penalizes the fill. AUC is the score-level AUC of separating the construction from the real removals with $H$ or $T$ as the third factor.}
\label{tab:copypaste}
\setlength{\tabcolsep}{3.5pt}
\begin{tabular}{lccccc}
\toprule
Construction & $\Delta\log S$ & $\Delta\log P$ & $\Delta\log H$ & AUC $H$ & AUC $T$ \\
\midrule
Copy-block & +0.259 & $-$0.150 & +0.238 & 0.660 & 0.595 \\
Duplicated object & +0.393 & +0.134 & +0.369 & 0.836 & 0.792 \\
Tiling & +0.656 & +0.308 & +0.424 & 0.920 & 0.906 \\
\bottomrule
\end{tabular}
\end{table}

\begin{table}[t]
\centering
\footnotesize
\caption{\textbf{Cost per image} on one NVIDIA RTX 4090 in full precision, the setting of every result in this paper. Times are averaged over 15 images and three repeats after one warm-up image, with CUDA synchronized around every stage and image decoding excluded. Both configurations run the shared encoder once and pass its features to the terms that need them.}
\label{tab:cost}
\setlength{\tabcolsep}{6pt}
\begin{tabular}{lcc}
\toprule
 & ERASCORE & ERASCORE-T \\
\midrule
Shared encoder pass (s) & 0.243 & 0.247 \\
$S$ (s) & 0.008 & 0.009 \\
$P$ (s) & 0.051 & 0.067 \\
$H$ or $T$ (s) & 2.624 & 0.006 \\
\midrule
Total (s) & \textbf{2.925} & \textbf{0.329} \\
Peak memory (GB) & 11.02 & 5.45 \\
Frozen networks & 4 & 2 \\
Parameters & 1.89\,B & 1.15\,B \\
\midrule
SRCC & 0.6739 & 0.6738 \\
\bottomrule
\end{tabular}
\end{table}

\subsection{Role of the Hallucination Term}
\label{sec:analysis_h}

$H$ adds little to average agreement, as expected for a term that acts on a narrow class of failures. Removing it lowers SRCC by only \V{0.0019} (Table~\ref{tab:terms}). We therefore test whether $H$ acts on the failures it targets. We examine natural hallucinations from strong removers and real scene pixels pasted into the mask, two settings in which patch-level evidence is least informative, and then ablate the parts of $H$.

\begin{table}[t]
\centering
\footnotesize
\caption{\textbf{Diagnostics of the hallucination term.} AUC separates the 53 flagged hallucinations of OmniPaint and ObjectClear from clean outputs of the same scenes (Sec.~\ref{sec:analysis_h}), with ties counted as 0.5, and the row for the other six methods gives its range over those methods. Clean FP is the share of clean outputs with $H>1.25$. Each row of the middle block removes one component of $H$, and $p$ compares it with the full $H$ ($^\dagger$Holm-adjusted over the component tests, otherwise exact McNemar). The last block counts candidates in duplicated objects and in flagged outputs.}
\label{tab:hdiag}
\setlength{\tabcolsep}{4pt}
\begin{tabular}{lccc}
\toprule
 & AUC & Clean FP & $p$ \\
\midrule
$S\cdot P$ & 0.610 & -- & -- \\
ERASCORE-T & 0.597 & -- & -- \\
ERASCORE & 0.635 & -- & -- \\
$H$ & \textbf{0.858} & 2.38\% & -- \\
$S\cdot P$, other six methods & 0.82--0.99 & -- & -- \\
\midrule
\multicolumn{4}{l}{\textit{$H$ without}} \\
\quad region proposals & 0.484 & -- & 0.002$^\dagger$ \\
\quad alignment $\rho$ & -- & 3.71\% & $<10^{-14\dagger}$ \\
\quad depth ($\rho$ and $\pi$) & -- & 3.06\% & $6.9\times10^{-5}$ \\
\quad protrusion $\pi$ & 0.855 & -- & 1.0$^\dagger$ \\
\midrule
\multicolumn{4}{l}{\textit{Appearance gate and protrusion}} \\
\quad $\varphi<0.1$, duplicated objects & \multicolumn{3}{l}{0.3\% of candidates} \\
\quad $\varphi<0.1$, flagged outputs & \multicolumn{3}{l}{0.2\% of candidates} \\
\quad $\varphi<0.3$ duplicates & \multicolumn{3}{l}{8 demoted, 0 promoted of 76} \\
\bottomrule
\end{tabular}
\end{table}

\paragraph{Natural hallucinations.}
$H$ penalizes the plausible hallucinations of strong removers, which $S\cdot P$ struggles to distinguish from clean fills. To study these failures, we take the outputs that both MLLM judges flag as hallucinated from OmniPaint and ObjectClear, the two methods that humans prefer most, and compare them with outputs of the same scenes for which neither judge reports hallucinated content, which we call clean. The judges only select these \V{53} outputs and never enter the human target. Here $S\cdot P$ separates hallucinated from clean outputs barely better than chance, with an area under the ROC curve (AUC) of \V{0.61}, whereas $H$ alone reaches \V{0.86} (Table~\ref{tab:hdiag}). For the other six methods, $S\cdot P$ already separates flagged hallucinations well. In the product, $H$ scales an $S\cdot P$ that barely separates these outputs, so the combined score gains less than $H$ alone suggests.

\paragraph{Pasted scene content.}
Fills made of real scene pixels are penalized by $S$ and $H$, and therefore by the product, even when $P$ prefers them. We construct three such fills on all \nCases{} samples, pasting an adjacent region (copy-block), copying an object from elsewhere in the image (duplicated object), or tiling a single background patch across the mask (Table~\ref{tab:copypaste}). $P$ assigns a lower penalty to the copy-block than to an average real removal, since the copied pixels retain realistic local appearance, whereas $S$ and $H$ penalize all three constructions. We attribute the rise in $S$ to the contextualized descriptors, since a pasted region in new surroundings no longer matches its source closely. Together, the penalties from $S$ and $H$ outweigh the preference of $P$ for the copy-block.

\paragraph{Component ablation.}
Region proposals, support-layer alignment, and the depth estimator each contribute to the targeted behavior, and removing any of them exposes the failure it was designed for. Without region proposals, with the mask itself as the only candidate, $H$ can no longer tell natural hallucinations from clean fills (AUC \V{0.48} instead of \V{0.86}), so the proposals are what make $H$ an object-level test (Table~\ref{tab:hdiag}). Support-layer alignment $\rho$ and depth protect valid background, and removing either one significantly raises the share of clean outputs that $H$ penalizes (Table~\ref{tab:hdiag}). Protrusion $\pi$ has the smallest measured effect, and removing it alone does not significantly change the AUC. We keep $\pi$ because it is the only signal that can amplify a plausible structure placed at the wrong depth, as the copied balls of Fig.~\ref{fig:qualitative} show. On the duplicated objects that resemble the scene most closely, it demotes \V{8} copies without promoting any. Gating $\pi$ by appearance does not silence it in practice, since $\varphi$ falls below 0.1 for at most \V{0.3\%} of the candidates in duplicated or hallucinated content (Table~\ref{tab:hdiag}).

\subsection{ERASCORE-T and Cost}
\label{sec:analysis_t}

ERASCORE-T keeps almost all of the average agreement at a small fraction of the cost. It reaches \srccT{} SRCC against \srccMain{} for the full metric, a difference a paired test does not detect. Without segmentation and depth, it runs about \speedup{} times faster with half the memory (Table~\ref{tab:cost}).

The full metric is slower, but $H$ is more sensitive to these hallucinations than $T$. Because $S\cdot P$ is shared, every difference between them comes from the third factor. $H$ separates every pasted construction from real removals better than $T$, most clearly on the copy-block (Table~\ref{tab:copypaste}), and on the natural hallucinations of strong removers the full metric separates flagged from clean outputs better than ERASCORE-T (Table~\ref{tab:hdiag}). We recommend ERASCORE-T for large-scale screening, where computational efficiency and average agreement are the main concerns. However, when detecting plausible hallucinations from strong removal models is a priority, we recommend the full metric for its greater sensitivity to such failures.

\subsection{Robustness}
\label{sec:analysis_robust}

\begin{table}[t]
\centering
\footnotesize
\caption{\textbf{Robustness} (808 samples). $\Delta$ and $p$ compare each ERASCORE variant with the adopted SAM ViT-H/DINOv2-Giant configuration, and the last row compares CFD with and without the four-way split. With SAM ViT-B, ERASCORE remains above all 29 baselines, including the strongest, Gemini-2.5-Pro at 0.6405 (paired $p=2.55\times10^{-4}$).}
\label{tab:robust}
\setlength{\tabcolsep}{4.5pt}
\begin{tabular}{lrrr}
\toprule
Configuration & SRCC & $\Delta$ & Paired $p$ \\
\midrule
\textbf{ERASCORE (ViT-H, Giant)} & \textbf{0.6739} & -- & -- \\
SAM proposals split into four & 0.6747 & +0.0008 & 0.214 \\
Drop 25\% of candidates & 0.6735 & $-$0.0004 & 0.777 \\
Drop 50\% of candidates & 0.6732 & $-$0.0008 & 0.621 \\
SAM ViT-B & 0.6720 & $-$0.0019 & 0.161 \\
DINOv2-Base & 0.6717 & $-$0.0022 & 0.296 \\
DINOv2-Large & 0.6701 & $-$0.0037 & 0.0268 \\
\midrule
CFD & 0.3195 & -- & -- \\
CFD proposals split into four & 0.3414 & +0.0218 & 0.000255 \\
\bottomrule
\end{tabular}
\end{table}

ERASCORE depends on segmentation, yet ranking agreement is robust to the tested segmentation perturbations. Splitting every SAM proposal into four pieces changes SRCC by less than \V{0.001}, while the same split moves CFD more than twenty times as much. Each fragment must still pass the appearance and geometry tests, so over-segmenting a clean fill rarely creates a penalty. Randomly dropping up to half of the candidates barely changes the ranking because the patch-level terms are unaffected. With the smaller SAM ViT-B, ERASCORE remains above all \nBaselines{} baselines (Table~\ref{tab:robust}).

\begin{table}[t]
\centering
\footnotesize
\caption{\textbf{Effect masks and large masks} (SRCC). The two OBER rows change only the mask that ERASCORE uses, with baselines and human reference fixed. Largest masks is the top quartile of mask area. The strongest baseline has the highest SRCC on each subset, and $p$ is its Holm-adjusted test against ERASCORE over the 29 baselines.}
\label{tab:subsets}
\setlength{\tabcolsep}{4pt}
\begin{tabular}{lrccc}
\toprule
Subset & $n$ & Ours & Strongest baseline & $p$ \\
\midrule
Object mask (OBER) & 201 & 0.725 & DreamSim, 0.706 & 0.218 \\
Object + effect mask & 201 & 0.762 & DreamSim, 0.706 & 0.012 \\
Largest masks & 202 & 0.714 & GPT-4o, 0.701 & 0.910 \\
\bottomrule
\end{tabular}
\end{table}

\paragraph{Effect masks and mask size.}
ERASCORE scores only the region of the target mask, but when a mask that also covers the effects of an object is available, nothing in the metric has to change. Using the object and effect masks of OBER raises agreement on the \V{201} samples that have them from \V{0.72} to \V{0.76} (Table~\ref{tab:subsets}). Very large masks remain harder. On the quarter of samples with the largest masks, ERASCORE is ahead of the strongest baseline only numerically (Table~\ref{tab:subsets}).

\section{Conclusion}
\label{sec:conclusion}

Our results show that, in reference-free evaluation of object removal, how the reference is built has a large effect on agreement with human judgments. ERASCORE takes its reference from the original image, lets it cover the entire unmasked scene, and keeps each scene element separate, and controlled ablations show that each of these choices improves agreement on its own. Reading that evidence in a semantic and a perceptual space lets the metric respond to both wrong content and poor rendering, and a bounded region term helps penalize plausible hallucinations of strong removal models that patch-level support often lets through. Without ground truth, task-specific training, or fitted weights, ERASCORE agrees with human rankings better than \nBaselines{} baselines, including MLLM judges, and it stays first under every human reference we tested. ERASCORE-T offers nearly the same agreement at a fraction of the cost. We will release the full evaluation code to support further work on removal evaluation.

\noindent\textbf{Limitations.} ERASCORE scores only the region of the target mask, so effects that lie entirely outside it are missed unless the mask covers them, and its lead on very large masks is only numerical (Table~\ref{tab:subsets}). Because $H$ is bounded, a severe but well-rendered hallucination can stop short of the bottom rank. The metric relies on four frozen networks whose errors can affect $H$, and its region branch accounts for most of the runtime. Finally, 453 of the \nCases{} samples carry a single human ranking, which makes the reference noisier on those samples, although ERASCORE remains first when the reference averages four rankings (Sec.~\ref{sec:results_mor}).

{
    \small
    \bibliographystyle{ieeenat_fullname}
    \bibliography{main}
}

\end{document}